\documentclass[conference]{IEEEtran}
\IEEEoverridecommandlockouts

\usepackage{cite}
\usepackage{amsmath,amssymb,amsfonts}
\usepackage{algorithmic}
\usepackage{graphicx}
\usepackage{textcomp}
\usepackage{xcolor}
\usepackage{colortbl}
\usepackage{multirow} 
\usepackage{booktabs}
\usepackage{authblk}
\usepackage{marvosym}
\usepackage{dcolumn}
\usepackage{hyperref}
\def\BibTeX{{\rm B\kern-.05em{\sc i\kern-.025em b}\kern-.08em
    T\kern-.1667em\lower.7ex\hbox{E}\kern-.125emX}}
\begin{document}

\title{PA-LLaVA: A Large Language-Vision Assistant for Human Pathology Image Understanding\\
}
\author[1]{Dawei Dai}
\author[1]{Yuanhui Zhang}
\author[1]{Long Xu}
\author[2]{Qianlan Yang} 
\author[2]{Xiaojing Shen}
\author[1]{Shuyin Xia}
\author[1,3]{Guoyin Wang}

\affil[1]{Chongqing Key Laboratory of Computational Intelligence,\protect\\ Chongqing University of Posts and Telecommunications, 400065, Chongqing, China.}
\affil[2]{Shanghai First Maternity and Infant Hospital, School of Medicine Tongji University, 200092, Shanghai, China}
\affil[3]{College of Computer and Information Science, Chongqing Normal University, 401331, Chongqing, China.}

\maketitle

\begin{abstract}
The previous advancements in pathology image understanding primarily involved developing models tailored to specific tasks. Recent studies has demonstrated that the large vision-language model can enhance the performance of various downstream tasks in medical image understanding. In this study, we developed a domain-specific large language-vision assistant (PA-LLaVA) for pathology image understanding. Specifically, (1) we first construct a human pathology image-text dataset by cleaning the public medical image-text data for domain-specific alignment; (2) Using the proposed image-text data, we first train a pathology language-image pretraining (PLIP) model as the specialized visual encoder for pathology image, and then we developed scale-invariant connector to avoid the information loss caused by image scaling; (3) We adopt two-stage learning to train PA-LLaVA, first stage for domain alignment, and second stage for end to end visual question \& answering (VQA) task. In experiments, we evaluate our PA-LLaVA on both supervised and zero-shot VQA datasets, our model achieved the best overall performance among multimodal models of similar scale. The ablation experiments also confirmed the effectiveness of our design. We posit that our PA-LLaVA model and the datasets presented in this work can promote research in field of computational pathology. All codes are available at: \textcolor{blue}{\href{https://github.com/ddw2AIGROUP2CQUPT/PA-LLaVA}{https://github.com/ddw2AIGROUP2CQUPT/PA-LLaVA}}
\end{abstract}


\begin{figure}[htbp]
	\centering
	\includegraphics[width=0.46\textwidth]{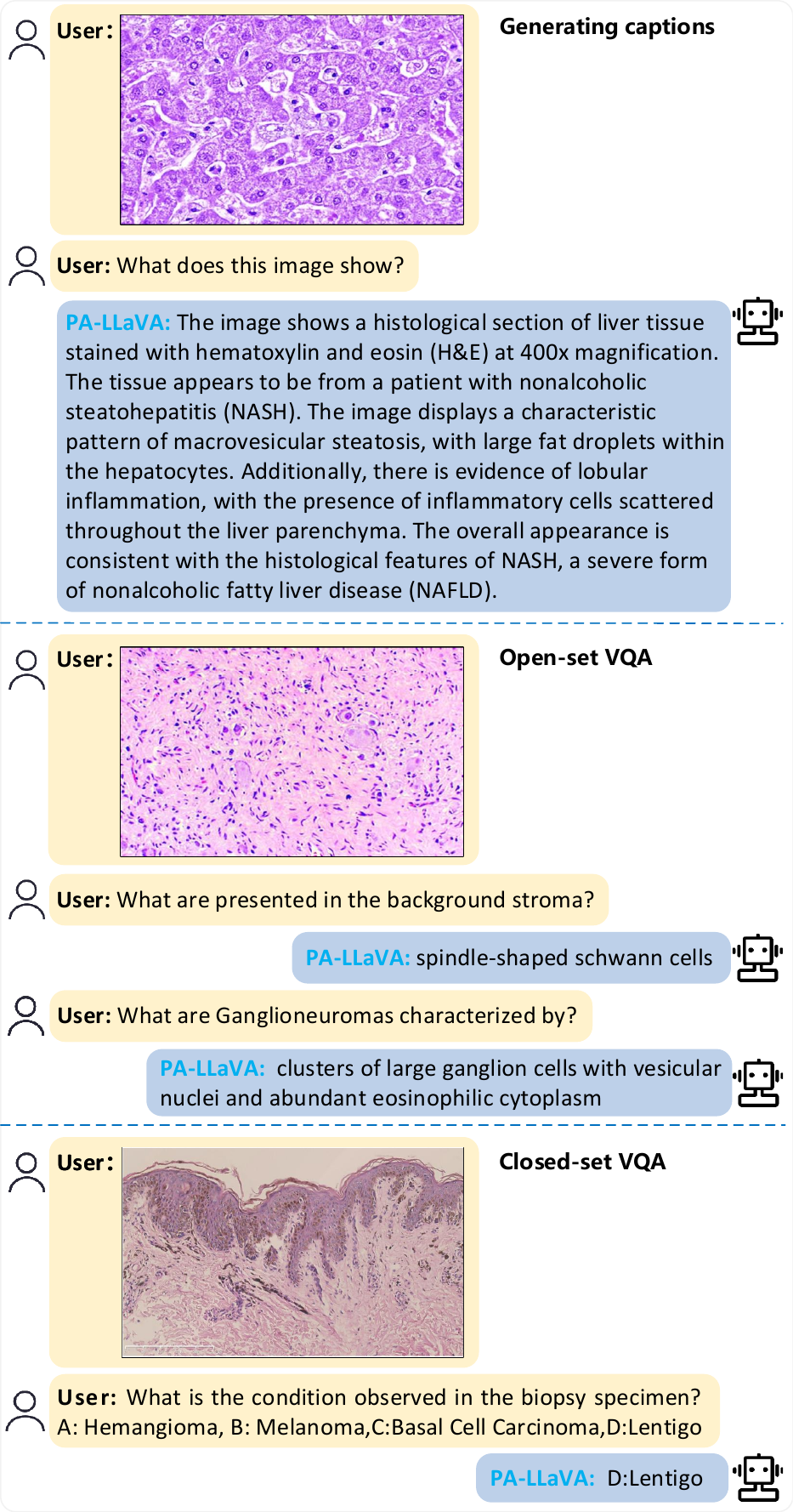}
	\caption{Demonstration of our PA-LLaVA. It is capable of answering various questions based on the pathology image. In this study, description generation is only an intermediate stage that requires high-quality pathological image-text data for fune-tuning to improve its quality. }
	\label{fig1}
\end{figure}

\section{Introduction}
The gold standard for diagnosing numerous diseases is tissue examination by a pathologist. This process involves meticulous analysis of tissue samples, typically using a microscope, to identify cellular abnormalities. Pathologists rely on extensive training and experience to interpret nuanced patterns and features that indicate various disease states, underscoring the critical role of pathology in accurate diagnosis and patient care. The application of artificial intelligence (AI) in pathology research stems from the time-consuming nature and subjective limitations of traditional pathology diagnostic methods. AI-based methods have demonstrated considerable advances in numerous tasks such as metastasis detection\cite{bejnordi2017diagnostic}, cancer subtyping\cite{lu2021data}, survival prediction\cite{courtiol2019deep,skrede2020deep,chen2022pan}, unknown primary origin site prediction\cite{lu2021ai,zhu2023accurate}, image search\cite{chen2022fast,wang2023retccl}, and prediction of molecular alterations\cite{kather2020pan,saldanha2023self}. However, recent advancements in this field primarily involve developing models tailored to specific tasks.

Generative pre-training has proven effective in leveraging the image-text data for self-supervised vision-language modeling, as evidenced by multimodal systems such as Large Language-Vision Assistant (LLaVA)\cite{liu2024visual}. By fine-tuning the large language model (LLM) to align multimodal inputs (image and text), the LLaVA demonstrates robust task completion capabilities across a variety of user-oriented vision-language tasks, including image understanding and reasoning. This advancement supports the development of versatile multimodal conversational assistants\cite{kuckreja2024geochat,bazi2024rs,cai2024vipllava,caffagni2024wiki}. However, LLAVA-based models that trianed in the general domain often underperform in the specialized field of medicine, providing incorrect answers or even complete hallucinations. \textbf{Therefore, constructing the specialized LLaVA models for specific domains is essential and meaningful.}

Inspired by the success of ChatGPT, numerous downstream tasks based on pathological images can be reframed as visual question-answering (VQA) task. Recently, several LLaVA-based models such as LLaVA-Med\cite{li2024llava}, PathChat\cite{lu2024multimodal}, and Quilt-LLaVA\cite{seyfioglu2024quilt} have gradually emerged in the field of medical image understanding. However, these models mainly focused on constructing specific medical datasets, with limited improvements at the model level, and many proprietary data are not open source, making reproducibility challenging. In this study, we developed a large language vision assistant model for pathology image understanding using only the public data. Studies have demonstrated that pre-training tailored to specific domains can be effective for the applications in biomedical vision-language integration. Therefore, we first trained a pathology language-image pre-training (PLIP) model to replace the general visual encoder. Subsequently, we designed a scale-invariant connector to avoid information loss caused by image scaling. \textbf{Our contributions are summarized as follows:}

\textbf{(1) Human Pathology Image-Text Instruction-following Data.} For the pathology domain alignment, we constructed a high-quality pathology image caption (two versions: 518,413 vs 827,401) by cleaning only the public medical image-text data. This caption describes the content of the image comprehensively. In the construction process, we extensively used large language models and other specialized models.

\textbf{(2) PA-LLaVA model.} Using self-constructed data, we developed an improved LLaVA-based model. We first trained a specialized PLIP model as the vision encoder rather than the general CLIP, demonstrating PLIP’s efficient representation of pathology images. Subsequently, we developed a scale-invariant connector to avoid information loss caused by image scaling. Our proposed PA-LLaVA achieved best overall performance on several public data.

\textbf{(3) Open-source.} To facilitate research in the field of pathology image understanding, we plan to release the following assets to the public: All the instruction-following data, PLIP model checkpoints, the first- and second-stage PA-LLaVA checkpoints, and the codebase for model training.

\begin{figure*}[htbp]
	\centering
	\includegraphics[width=1\textwidth]{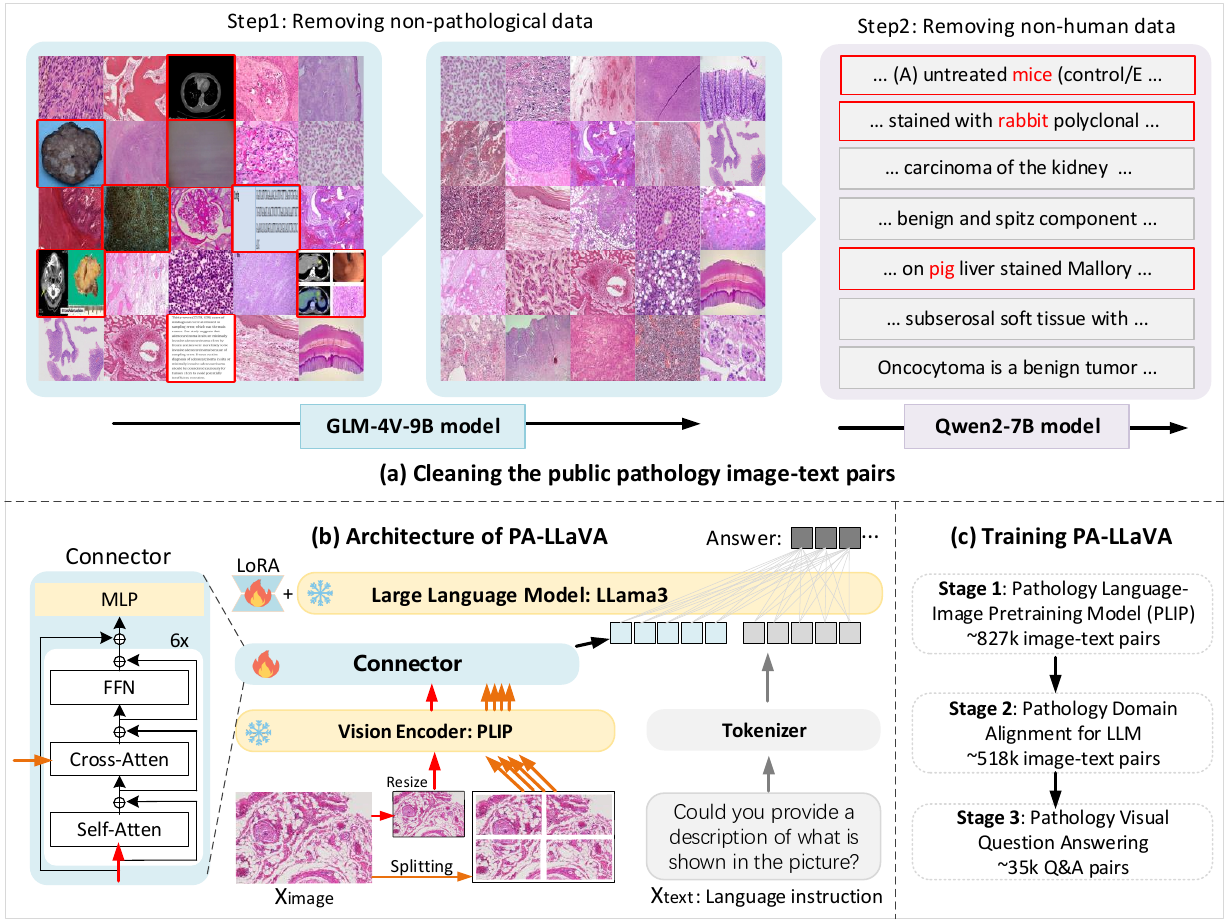}
	\caption{Overview of our PA-LLaVA. (a) Piplines of constructing the dataset of human pathology image-text pairs; (b) Architecture of our PA-LLaVA model; (c) Three-stages learning for PA-LLaVA.}
	\label{fig2}
\end{figure*}

\section{Related Work}
The previous advancements in pathology image understanding were achieved mainly through the framework of developing models tailored to the specific tasks. Liu et al.\cite{liu2024visual} pioneered the fully automated generation of multimodal language-image instruction-following data using large language models, and introduced a large language-vision assistant (LLaVA) trained end-to-end on such instruction-following data, aimed at general purpose of visual and language understanding. LLaVA sets new standards for efficiency and effectiveness in multimodal learning, and has been rapidly adopted in the medical image field. Unlike traditional methods, this new framework requires instruction-following image-text data. We summarize the research from the perspectives of medical image-text data and LLaVA-based medical model.

\subsection{Medical Image-Text Datasets.}

The difficulty in collecting medical datasets stems primarily from critical patient privacy concerns. Researchers have created multimodal medical datasets from public resources by using meticulous extraction methods. For instance, Lin et al. developed a biomedical dataset named PMC-OA\cite{lin2023pmc} from PubMed Central’s OpenAccess subset, which includes 1.6 million image-text pairs. Similarly, the PubMedVision\cite{chen2024huatuogpt} dataset was constructed from image-text pairs in PubMed and reformatted using GPT-4V. Ikezogwo et al. transformed unstructured videos from YouTube into vision-based instruction data and included other open-source data, such as Twitter and research papers, totaling one million image-text pairs to construct the Quilt-1M\cite{ikezogwo2024quilt} dataset. Furthermore, PMC-15M\cite{zhang2023biomedclip}, created by mining PubMed Central articles, utilizes figure caption pairs from these articles to compile the largest biomedical image-text dataset to date. Huang et al. constructed an OpenPath\cite{huang2023visual} dataset containing 208,414 pathology image-text pairs; however, only publicly provided model-processed feature vectors and not the original images. For medical VQA tasks, PMC-VQA\cite{zhang2023pmc} contains 227k question-answer pairs that cover various modalities and diseases based on images. For VQA datasets specifically targeting the field of pathology, PathVQA\cite{he2021towards} is a commonly used evaluation dataset, with each image associated with multiple questions categorized as either open-ended or closed-ended based on the nature of the answers.

\subsection{LLaVA for Medical Field.}

Li et al.\cite{li2024llava} first introduced the concept of LLaVA in the biomedical field, using GPT-4 to create a biomedical multimodal instruction-following dataset from PMC-15M. Subsequently, they trained a medical multimodal conversational assistant named LLaVA-Med. Lu et al.\cite{lu2024multimodal} adjusted the foundational visual encoder for pathology based on the LLaVA architecture to create PathChat, fine-tuning it on a self-created dataset containing 450,000 instruction pairs. On multiple-choice diagnostic questions and open-ended questions related to pathology, PathChat outperformed several multimodal visual-language AI assistants, including GPT-4V. Seyfioglu et al.\cite{seyfioglu2024quilt} also created an instruction-tuning dataset, QUILT-INSTRUCT, and an evaluation dataset, QUILT-VQA, containing 107,131 histopathology-specific question-answer pairs. They trained a domain-specific Quilt-LLaVA model that enhanced the performance on the QUILT-VQA and public histology VQA test datasets. Zhang et al.\cite{zhang2023biomedgpt} proposed BiomedGPT for various biomedical tasks that were pre-trained on 352,567 images, 183 million text tokens, 46,408 object label pairs, and 271,803 image-text pairs and achieved SOTA results in 16 out of 26 datasets across five clinically relevant tasks. Additionally, Van et al.\cite{van2023open} proposed an open-ended medical Visual Question Answering (CLIP-ViT w/GPT2 ) method based on pre-trained large language models, and introduced a novel approach particularly suited for small, domain-specific medical datasets. Li et al.\cite{li2023masked} adopted multimodal contrastive loss, masked language modeling loss, and image-text matching loss as pre-training objectives to align visual and textual information, and achieved SOTA performance on three public medical VQA datasets.

\section{Methodology}
An overview of our PA-LLaVA model involving a three-stage learning process is as illustrated in \textbf{Fig} \ref{fig2} (c). Stage 1 trains a pathology language-image pre-training (PLIP) model to perform the role of the visual encoder. Stage 2 achieves the pathology domain alignment, while stage 3 involves VQA instruction fine-tuning. The data used in the three stages can be classified into two categories (some illustrations are presented in \textbf{Fig} \ref{fig3}): (1) Image-Caption Data: consists of pathology images paired with natural language descriptions accurately describing the content of pathology images. (2) VQA Data: This involved questions and answers regarding the content of pathology images. All the training data were obtained from publicly available medical image-text datasets. To obtain the human pathology image-text data, we used publicly available models for data cleaning. To further verify the capability of our  PA-LLaVA on the zero-shot task, we transformed several pathology image classification datasets into VQA data to evaluate our PA-LLaVA.

\begin{figure*}[htbp]
	\centering
	\includegraphics[width=1\textwidth]{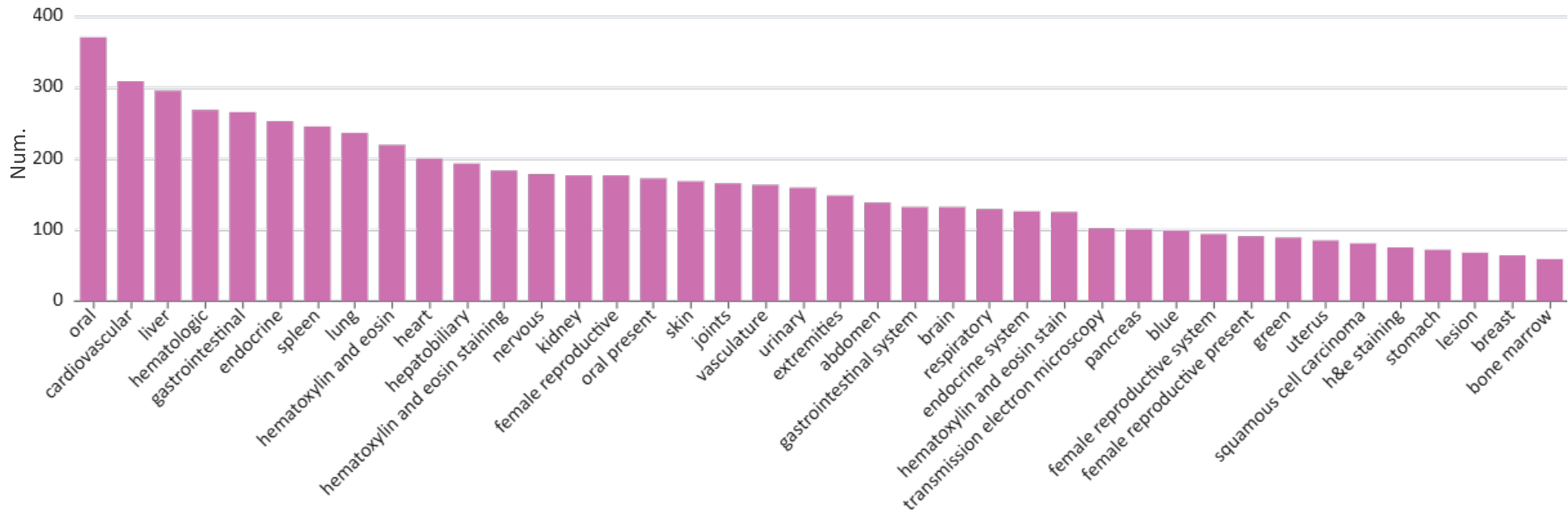}
	\caption{Frequencies of noun words in our VQA training data.}
	\label{fig4}
\end{figure*}

\subsection{Data Construction}

\subsubsection{Pathology Image-Caption Data}

The raw data were obtained from three datasets: Quilt-1M, PMC-OA, and PubMedVision Alignment, totaling 1,409,058 image-text pairs (named \textbf{“PCaption-C” }). (1) \textbf{Quilt-1M:} This contains one million image-text pairs from YouTube, Twitter, research papers, and the general Internet; it addresses the scarcity of similar data in the medical domain, particularly in pathology. (2) \textbf{PMC-OA:} It includes 1.6 million image-caption pairs collected from the open-access subset of PubMed Central. (3) \textbf{PubMedVision-Alignment:} We selected a sub-dataset of  microscopic images; this dataset commonly used in pathology-related research, totaling 132,973 image-description pairs.

\begin{figure}[htbp]
	\centering
	\includegraphics[width=0.48\textwidth]{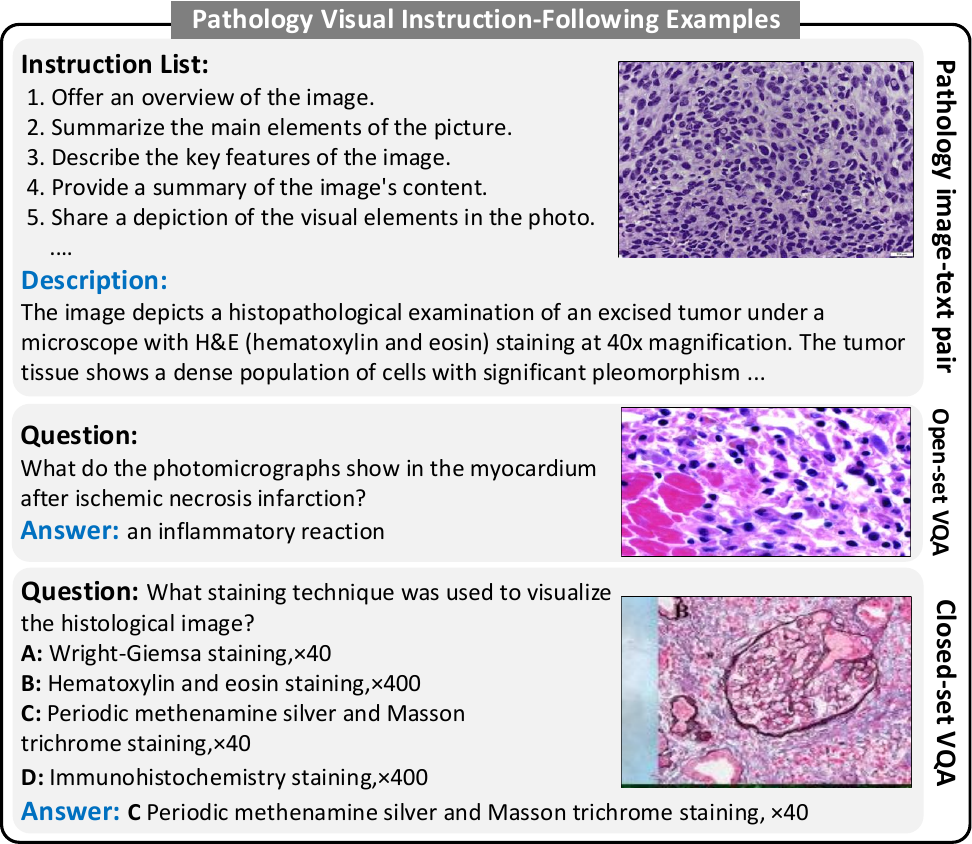}
	\caption{Illustrations of our instruction-following data.}
	\label{fig3}
\end{figure}

However, these datasets contain substantial amounts of data unrelated to human pathology. To obtain the human pathology image-text data, we performed two cleaning processes on the raw data, as illustrated in \textbf{Fig} \ref{fig2} (a): \textbf{(1) Removing non-pathological images} We used the GLM-4V-9B\cite{team2024chatglm} multimodal language model, which has strong visual understanding capabilities, to process the original image-text pairs. By inputting appropriate prompts into the model to determine whether an image is non-pathological; if the model returns yes, we remove the image-text pair.  \textbf{(2) Removing non-human pathology data:} We employed the Qwen2-7B-Instruct\cite{yang2024qwen2} model and input appropriate prompts to it to detect text descriptions. This step determines whether text descriptions involve non-human organisms. We deleted image-text pairs that included such references. After processing, the Quilt-1M and PMC-OA datasets were reduced to 584,195 and 110,233 pairs, respectively. Combined with the PubMedVision-Alignment subset, this results in approximately 827,401 refined pathology image-description pairs (named \textbf{“PCaption-0.8M” for stage 1}). Additionally, we excluded image-text pairs with textual descriptions of fewer than 20 words. Ultimately, we obtained 518,413 image-text pairs (named \textbf{“PCaption-0.5M” for stage 2}) for the aligned training dataset.

\subsubsection{VAQ Data for Stage 3}

This type of training data was obtained from PathVQA \cite{he2021towards} and PMC-VQA \cite{zhang2023pmc}. \textbf{PathVQA}: This is a high-quality dataset specifically designed for visual question-answering (VQA) tasks in pathology; It aggregates 4,998 pathology images, generating 19,755 question-answer pairs. \textbf{PMC-VQA}: This was a multimodal dataset containing 227k VQA questions and encompassing 149k medical images covering various medical modalities or diseases; After the cleaning, we obtained 15,788 question-answer pairs related to human pathology. Lastly, we combined PathVQA and Human pathology data obtained from PMC-VQA, thereby constructing a training dataset of 35543 question-answer pairs as the training data. As illustrated in \textbf{Fig} \ref{fig4}, we count the frequency of noun words in the text, which can reflect the domain distribution of the samples. Additionally, for the test dataset (named “Test” and “Test-Clean”) in PMC-VQA, we also perform the cleaning operations to obtain the human pathology image-text data: \textbf{“HT”} and \textbf{“HT-clean”}.

\subsubsection{VQA Data for Zero-Shot Test}

We selected three pathology image classification datasets (\textbf{ICIAR 2018 BACH\cite{aresta2019bach}, OSCC\cite{rahman2019histopathological} and ColonPath\footnote{https://medfm2023.grand-challenge.org/datasets}}) and transformed them into the closed-set VQA task. \textbf{ICIAR 2018 BACH} is a Breast Cancer Histology (BACH) dataset that classifies 400 H\&E-stained breast histology microscope images evenly into four categories based on the type of breast cancer: normal, benign, in situ, and invasive carcinoma. \textbf{OSCC} is the initial component of the histopathological imaging database for oral cancer analysis and includes 528 images categorized into normal oral epithelium and oral squamous cells. \textbf{ColonPath} is a binary classification dataset for lesion cells in colonic tissues. We used its validation set, which contains 5,355 tissue images with 2,727 positive samples and 2,628 negative samples, to evaluate the model. \textbf{The following questions were used to evaluate the models for the three datasets (How to design the prompt is still an open problem):}

“What choice best describes this breast tissue? Just give your choice: A:Normal tissue  B: Benign tumors C: In situ cancer D: Invasive cancer”; “What choice best describes this oral epithelium tissue? Just give your choice: A:Normal oral epithelium  B:Oral squamous cell carcinoma”; “What choice best describes this colon tissue? Just give your choice: A:Normal tissue B:Tumor tissue”

\renewcommand{\arraystretch}{1.2} 
\begin{table*}[h!]
	\centering
	\normalsize
	\caption{Comparison with prior state-of-the-art supervised methods.}
	\setlength{\tabcolsep}{.34cm}{
		\begin{tabular}{>{\centering\arraybackslash}m{3.5cm}|c|c|c|c|c|c}
			\toprule[1pt]
			\multirow{3}{*}{\textbf{Model}} & \multicolumn{3}{c|}{\textbf{PathVQA}} & \multicolumn{3}{c}{\textbf{PMC-VQA (Closed-set)}} \\
			\cline{2-7}
			& \multicolumn{1}{c|}{\textbf{Closed}} & \multicolumn{1}{c|}{\textbf{Open}} & \multicolumn{1}{c|}{\textbf{Overall}} & \multicolumn{1}{c|}{\textbf{HT}} & \multicolumn{1}{c|}{\textbf{HT-Clean}} & \multicolumn{1}{c}{\textbf{Overall}} \\ 
			& \multicolumn{1}{c|}{Acc.$\uparrow$} & \multicolumn{1}{c|}{Acc.$\uparrow$ / Rec.$\uparrow$} & \multicolumn{1}{c|}{Acc.$\uparrow$ / Rec.$\uparrow$} & \multicolumn{1}{c|}{Acc.$\uparrow$} & \multicolumn{1}{c|}{Acc.$\uparrow$} & \multicolumn{1}{c}{Acc.$\uparrow$}\\ 
			\toprule[1pt]
			LLaVA\cite{liu2024visual} & 63.20 & - / 7.74 & - /35.56 & 26.36 & 26.83 & 26.60 \\ 
			Quilt-LLaVA\cite{seyfioglu2024quilt} & 58.68 & - / 15.06 & - / 36.98 & 31.91 & 34.76 & 33.34 \\ 
			M2I2\cite{li2023self} & 88.00 & 36.30 / -  & 62.20 / - & - & - & - \\ 
			MUMC\cite{li2023masked} & 90.40 & 39.00 / -  & 65.10 / - & - & - & -\\ 
			BiomedGPT\cite{zhang2023biomedgpt} & 88.0 & 28.00 / -  & 58.10 / - & - & - & - \\ 
			CLIP-ViT w/GPT2\cite{van2023open} & 87.00 & \textbf{40.00} / -  & 63.60 / - & - & - & -  \\ 
			LLaVA-Med\cite{li2024llava} & 91.09 & - / 39.60 & - / 65.43 & 19.68 & 23.78 & 21.73 \\ 
			\rowcolor{gray!20} PA-LLaVA(Ours) & \textbf{92.51} & 38.04 / \textbf{42.20} & \textbf{65.36} / \textbf{67.43} & \textbf{38.93} & \textbf{40.85} & \textbf{39.89} \\ 
			\toprule[1pt]
		\end{tabular}
	}
	\label{table1}
\end{table*}

\subsection{Architecture of PA-LLaVA}

As illustrated in \textbf{Fig} \ref{fig2} (b), our PA-LLaVA consists of a vision encoder to extract the features of the pathology images; a connector that maps the tokens of the image to a specific number and dimension; and a LLM to output the answer. For our PA-LLaVA model, we first obtained the initial representation of the input pathology image using a PLIP model. Subsequently, the visual representation is encoded using a learnable connector, combined with tokenized textual queries, and fed into a LLM to generate the expected response. 

In the original LLaVA model, the visual encoder is typically a pre-trained encoder trained on a generic domain dataset, which requires the input images to be scaled to a fixed size. Introducing LLaVA into computational pathology presents two major challenges: (1) The general pre-trained model may have weak cross-domain generalization when transferred to the medical domain. To address this, we trained a PLIP model using our self-constructed pathology image-text dataset as the image encoder to enhance the feature representation capability of the pathology images. We adopted the same architecture as BLIP\cite{li2022blip} for our PLIP model. (2) For pathological images, scaling can lead to changes or loss of detailed features. Consequently, we employed a dynamic high-resolution strategy to retain the original image size. Pathological images typically contain comprehensive details and complex structures crucial for accurate diagnosis and analysis.

\subsection{Three-Stages Learning for PA-LLaVA}

\textbf{Pathological Image-Text Pretraining Model.} We introduced two widely used loss functions to train the PLIP model: Image-Text Contrastive (ITC) loss\cite{DAI2024102535} and Image-Text Matching (ITM) loss\cite{DAI2024102535}. The ITC loss aligns the two modalities of pathological images and text in the embedding space. The ITM loss function promotes fine-grained fusion between the two modalities, essentially representing a classification task where the model predicts whether the input image and text are matching pairs. Our PLIP model was trained using the PCaption-0.8M dataset. 

\textbf{Pathological Image-Text Cross-Domain Alignment for LLM.} Same with LLaVA, we also employed LM loss\cite{liu2024visual} to train our PA-LLaVA. This training stage aligns the pathological image and semantics for the LLM. Specifically, the PA-LLaVA learns to generate descriptions of pathological images. We designed a series of questions related to image content and combined them with PCaption-0.5M, resulting in 518,413 question-answer pairs. For each pathological image, questions regarding the image content were randomly sampled from the designed question set. By guiding the model to generate descriptions, we established an initial alignment of the model with pathological images and semantic descriptions, laying the foundation for VQA fine-tuning. During the training, we froze the visual encoder and LLM, only training the connectors and LoRA that added to LLM. In the traditional LLaVA, only connectors are trained at this stage. However, we found that fine-tuning both connector and LoRA of the LLM can obtain the lower alignment errors.

\textbf{VQA Instruction Learning.} We used multimodal visual question-answer (VQA) instruction-following data, including PathVQA and a sub-dataset of PMC-VQA to fine-tune the LLM and connector parts. The VQA dataset, containing 35,543 pairs, introduced greater diversity in questions and answers. This enhanced the ability of the PA-LLaVA model to respond accurately to various types of pathology-related instructions. Ultimately, our model could answer specific pathological questions and performed excellently on some public pathological test datasets.

\section{Experiments}

\subsection{Evaluation Metrics}
For the supervised task, we reported the accuracy (Acc.) /percentage of ground-truth tokens appearing in the generated sequences (closed-set questions); For open-ended questions, recall (Rec.) was used to measure the ratio of the ground-truth tokens that appeared in the generated sequences. For zero-shot task (classification task), we employed the metrics such as Acc., Rec., and Precision (Pre.) to further evaluate the model's ability to distinguish each category.

\renewcommand{\arraystretch}{1.2} 
\begin{table*}[h]
	\centering
	\normalsize
	\caption{ Comparisons on Zero-shot tasks.}
	\setlength{\tabcolsep}{.22cm}{
		\begin{tabular}{>{\centering\arraybackslash}m{2.6cm}|ccc|ccc|ccc|ccc}
			\toprule[1pt]
			\multirow{2}{*}{\textbf{Model}} & \multicolumn{3}{c|}{\textbf{BACH}} & \multicolumn{3}{c|}{\textbf{OSCC}} & \multicolumn{3}{c|}{\textbf{ColonPath}} & \multicolumn{3}{c}{\textbf{Overall}} \\ 
			& Acc.$\uparrow$ & Rec.$\uparrow$ & Pre.$\uparrow$ & Acc.$\uparrow$ & Rec.$\uparrow$ & Pre.$\uparrow$ & Acc.$\uparrow$ & Rec.$\uparrow$ & Pre.$\uparrow$ & Acc.$\uparrow$ & Rec.$\uparrow$ & Pre.$\uparrow$ \\ 
			\toprule[1pt]
			LLaVA-Med\cite{li2024llava} & 25.50 & 25.50 & 35.43 & 21.97 & 53.08 & 58.88 & 58.90 & 58.26 & 66.06 & 35.46 & 45.61 & 53.46\\
			Quilt-LLaVA\cite{seyfioglu2024quilt} & 33.75 & 33.75 & 48.43 & 28.60 & 56.62 & 58.71 & 74.36 & 74.72 & 78.78 & 45.57 & 55.03 & 61.97\\
			GPT4-o & 35.50 & 35.50 & 44.72 & \textbf{82.39} & 52.23 & \textbf{60.64} & 56.12 & 56.91 & 75.50 & 58.00 & 48.21 & 60.29\\
			\rowcolor{gray!20} PA-LLaVA(Ours) & \textbf{41.75} & \textbf{41.75} & \textbf{50.04} & 67.23 & \textbf{66.86} & 60.00 & \textbf{79.96} & \textbf{80.01} & \textbf{80.05} & \textbf{62.98} & \textbf{62.87} & \textbf{63.36} \\
			\toprule[1pt]
		\end{tabular}
	}
	\label{table2}
\end{table*}

\subsection{Implementation details}

\textbf{PLIP:} We implemented training using the PyTorch framework and trained the model for 30 epochs on 4×NVIDIA A100 GPUs. We used a batch size of 4 × 48. We employed the AdamW  optimizer with a weight decay of 0.01. The learning rate was initialized to 1e-5, warmed up to 1e-4 in 1,000 steps, and decayed at fixed intervals to 5e-5 over 30 epochs. Furthermore, we used randomly cropped images with a resolution of 224×224 as input. The text was truncated to a length of 100.

\textbf{PA-LLaVA:} We implemented the PA-LLaVA model using the Xtuner toolkit and trained it on 16 × NVIDIA A100 GPUs. Our training process is divided into two stages: alignment phase and instruction fine-tuning phase. (1) For stage 1, we set the gradient accumulation steps to 6, and the batch size was set to 16 × 6 × 6; The learning rate was linearly increased from the initial value to 1e-4 and then gradually decayed to 0 using the cosine annealing strategy. This phase of training lasted for 9 epochs. (2) For stage 2, the batch size was 16 × 6 × 4; The learning rate of the connector module was linearly increased from 1e-5 to 1e-4 and then cosine decayed to 1e-6; Meanwhile, the learning rate of the LLM’s LoRA gradually increased to 2e-4 and finally also cosine decayed to 1e-6; This training was conducted for 12 epochs. We used the AdamW optimizer and used mixed precision training to improve computational efficiency and save memory.

\begin{table*}[!h]
	\centering
	\normalsize
	\caption{Ablation experiments. (PC-0.5M $\rightarrow$ PCaption-0.5M, PC-C $\rightarrow$ PCaption-C, Conn. $\rightarrow$ Connector)}
	\setlength{\tabcolsep}{.06cm}{
		\begin{tabular}{c|c|c|c|c|c|>{\centering\arraybackslash}p{1.1cm}>{\centering\arraybackslash}p{1.1cm}|>{\centering\arraybackslash}p{1.1cm}>{\centering\arraybackslash}p{1.1cm}>{\centering\arraybackslash}p{1.1cm}|>{\centering\arraybackslash}p{1.1cm}>{\centering\arraybackslash}p{1.1cm}>{\centering\arraybackslash}p{1.1cm}}
			\toprule[1pt]
			 \multirow{2}{*}{\textbf{Model}} & \multirow{2}{*}{\textbf{PC-0.5M}} & \multirow{2}{*}{\textbf{PC-C}} & \multirow{2}{*}{\textbf{CLIP}} & \multirow{2}{*}{\textbf{PLIP}} & \multirow{2}{*}{\textbf{Conn.}} & \multicolumn{2}{c|}{\textbf{PathVQA}}  & \multicolumn{3}{c|}{\textbf{PMC-VQA (HT-clean)}} & \multicolumn{3}{c}{\textbf{Table.II (Overall)}} \\ 
			& & & & & & Acc.$\uparrow$ & Rec.$\uparrow$ & Acc.$\uparrow$ & Rec.$\uparrow$ & Pre.$\uparrow$ & Acc.$\uparrow$ & Rec.$\uparrow$ & Pre.$\uparrow$ \\ 
			\toprule[1pt]
			PA-LLaVA* & \checkmark &  & \checkmark &  &  & 61.56 & 63.71 & 39.63 & 36.13 & 35.80 & 37.13 & 43.87 & 43.46 \\
			PA-LLaVA\dag &  & \checkmark &  & \checkmark & \checkmark & 64.64 & 66.66  & 34.76 & 31.73 & 31.58 & 60.36 & 49.99 & \textbf{67.48} \\
			PA-LLaVA\ddag & \checkmark &  & \checkmark &  & \checkmark & 64.61 & 66.76  & 40.85 & \textbf{38.43} & 37.20 & 48.29 & 54.45 & 54.07\\
			\rowcolor{gray!20} PA-LLaVA (Ours) & \checkmark &  &  & \checkmark & \checkmark & \textbf{65.36} & \textbf{67.43} & \textbf{40.85} & 37.61 & \textbf{39.04} & \textbf{62.98} & \textbf{62.87} & 63.36 \\
			\toprule[1pt]
		\end{tabular}
	}
	\label{table3}
\end{table*}

\subsection{Comparisons with SOTA}
We compared our PA-LLaVA with popular methods, including the general domain LLaVA\cite{liu2024visual}, and large language visual models of the medical domain (LLaVA-Med\cite{li2024llava} and Quilt-LLaVA\cite{seyfioglu2024quilt}), and the previous SOTA methods on pathology VQA task, with results shown in \textbf{Table} \ref{table1}. We can make the main observations as follows: 

First, our PA-LLaVA significantly outperformed the general domains of LLaVA. The main reasons can be noted as follows: (1) the LLM LLama3 performed better than that of Vicuna employed by LLaVA; (2) the initialization of the vision encoder from our PLIP model, pre-trained on pathological image-text data, surpasses that of general-domain CLIP\cite{radford2021learning}; (3) our PA-LLaVA has undergone cross-domain alignment of pathological images, text, and fine-tuning of visual question answering data.

Second, our PA-LLaVA performed better than the supervised SOTA on both closed-set and open-set VQA on PathVQA and PMC-VQA datasets. The main reasons are as follows: (1) Unlike other SOTA methods, where visual encoders require resizing medical images to a fixed size, the designs of our visual encoder and connector preserved the original size of pathological images, avoiding information loss caused by image scaling; and (2) the features extracted from the PLIP are more efficient than those of the general encoder used in previous SOTA models.

Third, we further evaluated our PA-LLaVA on the zero-shot task. The comparisons are presented in \textbf{Table} \ref{table2}. We can make the main observations that our PA-LLaVA significantly outperformed both the general and medical domain models. The main  because can be noted that our visual encoder and connector provide a more efficient feature representation for the pathology image than the original structure of LLaVA used in other LLaVA-based medical models. Additionally, compared to these models, the data used in our PA-LLaVA only focuses on the field of pathology. As shown in Fig \ref{fig4}, owing to the unbalanced distribution of pathological images used for the VQA task, our PA-LLaVA can exhibit different performance on different organs of pathological image.

\subsection{Ablation Studies}

To verify the effectiveness of each module—including our instruction data, domain-specific visual encoder, and connector. We reported the performance of the different PA-LLaVA variants in \textbf{Table} \ref{table3}. PA-LLaVA represents the proposed model; PA-LLaVA*, PA-LLaVA\dag and PA-LLaVA\ddag represent variants using different modules.

The main observations are as follows: (1) Our PA-LLaVA trained on PCaption-0.5M significantly outperformed the PA-LLaVA\dag that trained on PCaption-C data. This indicates that our data-cleaning process was effective for downstream tasks. (2) When our specialized visual encoder PLIP in PA-LLaVA was replaced with the general visual encoder CLIP, while keeping all other conditions unchanged (PA-LLaVA vs PA-LLaVA\ddag ), the performance degraded, especially on zero-shot tasks. This indicates that our specialized PLIP model can provides more effective feature representations for the pathology images. (3) Building on the previous step, when we replaced the connector part with an MLP structure (PA-LLaVA* vs PA-LLaVA\ddag ), which reverted to the original architecture of LLaVA, the model performance was further degraded, which confirmed the effectiveness of our proposed connector.

\subsection{Alignment Evaluation for LLM}

This stage aligns the embeddings of pathology image with their corresponding embeddings of text. After this training stage, our PA-LLaVA model generated descriptions of pathology images based on specific prompts. To verify performance, we compared our PA-LLaVA with Quilt-LLaVA and LLaVA-Med. Specifically, we leverage GPT-4 to evaluate the semantic similarity of the responses and captions, and give an overall score on a scale of 1 to 10, where a higher score indicates better overall performance. As listed in \textbf{Table} \ref{table4}, the quality of description generated by our proposed PA-LLaVA is closer to LLaVA-Med, significantly better than Quilt-LLaVA. Typically, the better the model aligns during this stage, the better its understanding of pathology images, and the more likely it is to perform downstream tasks (see PA-LLaVA vs PA-LLaVA\dag in \textbf{Table} \ref{table3}). However, this stage typically requires a large-scale dataset of images and content descriptions, which is mainly obtained on the Internet, often resulting in lower-quality text annotations. Thereby, constructing high-quality pathology image-text data are meaningful to our PA-LLaVA. 

\begin{table}[h!]
	\centering
	\small
	\caption{Anlignment Evaluation}
	\setlength{\tabcolsep}{.9cm}{
		\begin{tabular}{c|c}
			\toprule[1pt]
			\multicolumn{1}{c|}{\textbf{Model}} & \multicolumn{1}{c}{\textbf{GPT4 Score}} \\ 
			\toprule[1pt]
			Quilt-LLaVA\cite{seyfioglu2024quilt} & 2.32 \\
			LLaVA-Med\cite{li2024llava} & \textbf{4.77} \\
			\rowcolor{gray!20} PA-LLaVA(Ours) & 4.55 \\
			\toprule[1pt]
		\end{tabular}
	}
	\label{table4}
\end{table}

\section{Discusions}

\textbf{Limitations.} The quality of our pathology image-text pairs used for domain alignment still requires improvement. This is because domain alignment requires a sufficiently large sample size. Since this type of training data is mainly sourced from the internet, ensuring the quality of the images and texts is challenging. In this study, we performed the data cleaning on the public data, and constructed a dataset of human pathology image-text pairs. However, we cannot ensure their quality or correctness due to the highly specialized nature of interpreting pathology images. Therefore, improving the high-quality descriptions of the pathology images is beneficial for downstream tasks. Additionally, more balanced distribution of pathological VQA data can further improve the comprehensive performance of our PA-LLaVA.

\textbf{Conclusions.} (1) We constructed human pathology image-text instruction-following data by cleaning only the public data for the computational pathology field. However, we posit that the quality of these image texts can be further improved. (2) We developed an improved LLaVA-based model for understanding pathology images, where a PLIP model was used as a vision encoder and a scale-invariant connector was designed to avoid information loss caused by image scaling. To test its performance, we evaluated our PA-LLaVA on both supervised and zero-shot VQA datasets, where it achieved the best overall performance. We posit that the PA-LLaVA model and multimodal datasets presented in this study can promote research in the field of computational pathology.

\bibliographystyle{IEEEtran}
\bibliography{IEEEabrv,ref}

\end{document}